\documentclass[10pt,twocolumn,letterpaper]{article}
\usepackage[final]{cvpr}      
\usepackage{subfloat}
\usepackage{makecell,rotating,multirow,diagbox}

\newtheorem{remark}{Remark}[section]
\newtheorem{defn}{Definition}[section]
\usepackage[table]{xcolor}   
\definecolor{bestcell}{HTML}{FFD9D9}
\definecolor{secondcell}{HTML}{DEEAFB}
\definecolor{cvprblue}{rgb}{0.21,0.49,0.74}
\usepackage[pagebackref,breaklinks,colorlinks,allcolors=cvprblue]{hyperref}
\def\paperID{15784} 
\def\confName{CVPR}
\def\confYear{2026}
\title{MAUGIF: Mechanism-Aware Unsupervised General Image Fusion   via Dual Cross-Image Autoencoders}
\author{
Kunjing Yang$^{1,\dagger}$ \quad
Zhiwei Wang$^{2,3,\dagger}$ \quad
Minru Bai$^{1,*}$\\[1ex]
$^{1}$School of Mathematics, Hunan University, Changsha, Hunan, 410082, P. R. China\\
$^{2}$College of Information Engineering, Zhejiang University of Technology, Hangzhou, 310010, China\\
$^{3}$Department of Electrical and Electronic Engineering, The University of Hong Kong, Hong Kong, China\\[0.5ex]
{\tt\small kunjing-yang@hnu.edu.cn \quad wzw@zjut.edu.cn \quad minru-bai@hnu.edu.cn}\\[1ex]
$^{\dagger}$Equal contribution. \quad $^{*}$Corresponding author.
}
\begin{document}
\maketitle
\begin{abstract}
	Image fusion aims  to integrate  structural and complementary information from multi-source images. However, existing fusion methods are often either highly task-specific, or general  frameworks that apply uniform strategies across diverse tasks, ignoring  their distinct fusion mechanisms. To address this issue,  we propose a mechanism-aware unsupervised general image fusion (MAUGIF) method based on  dual cross-image autoencoders. Initially, we introduce a classification of additive and multiplicative fusion according to the inherent  mechanisms of different fusion tasks. Then, dual encoders map source images into a shared latent space, capturing common content while isolating modality-specific details. During the decoding phase, dual decoders act as feature injectors, selectively reintegrating the unique characteristics of each modality into the shared content for reconstruction. The modality-specific features are injected  into the source image in the fusion process, generating the fused image that integrates information from both modalities. The architecture of  decoders varies according to their fusion mechanisms, enhancing both performance and interpretability. Extensive experiments are conducted on diverse fusion tasks to validate the effectiveness and generalization ability of our method. The code is available at 
    \href{https://github.com/warren-wzw/MAUGIF.git}{https://github.com/warren-wzw/MAUGIF.git}.
\end{abstract}

\section{Introduction}
Image fusion aims to integrate complementary information from multiple source images into a single output that is more informative, visually interpretable, and suitable for downstream tasks such as object detection \cite{He2023}, segmentation \cite{Wang2025DiFusionSegDS}, and remote sensing analysis \cite{Hong2018}.
General image fusion refers to designing a unified model or framework that can be applied to various types of image fusion tasks. These tasks include hyperspectral-multispectral image  fusion (HMF), visible-infrared image fusion (VIF), multi-focus image fusion (MFF), and medical image fusion (MEF). The HMF task  seeks to fuse a low-spatial-resolution hyperspectral image (HSI) with a high-spatial-resolution multispectral image (MSI) to produce a high-spatial-resolution HSI \cite{Hysure}. The VIF integrates thermal radiation data from infrared sensors with rich texture and color information from visible cameras. This results in fused images that simultaneously highlight salient objects  and retain visual fidelity \cite{Ma2021GANMcCAG}. The MFF task aims at generating  an all-in-focus image from a series of partially focused images \cite{Quan2025}. The purpose of MEF is to  display the fine structure and functional metabolism of the lesion  simultaneously, thereby assisting doctors in making more accurate diagnoses and assessment \cite{HE2025102666}.
\begin{figure}[!t]
	\vspace{-4pt}
	\centering
	\!\!\includegraphics[width=6.3cm]{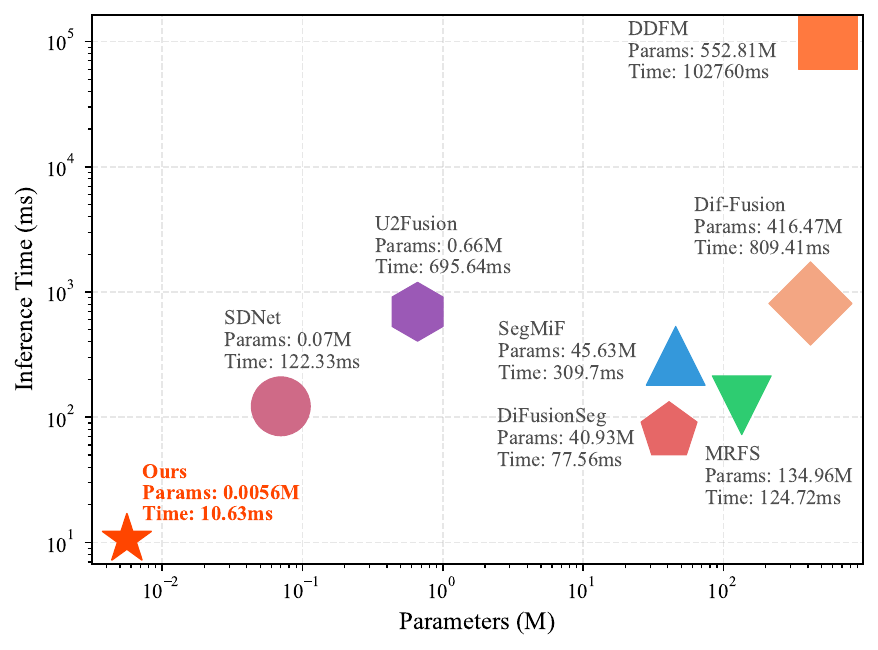}
    
    \vspace{-10pt}
	\caption{{\small The number of model parameters and inference time of different fusion methods.}}
	\label{f11}
    \vspace{-15pt}
\end{figure}

Existing image fusion methods can be broadly categorized into task-specific and general-purpose.  Task-specific methods are specially designed for a particular type of image fusion task. These methods typically leverage prior image information such as sparsity \cite{OTD} and structural sparsity \cite{S4}, or deep image features \cite{ZSL} to enhance fusion performance \cite{Zhang2023}. However, these methods often exhibit limited generalization capability across different fusion tasks. In contrast,
general image fusion methods aim to adopt a uniform fusion strategy across tasks \cite{Xu2020U2FusionAU}. These methods generally use general features including  frequency coefficients \cite{Wang2024a} and deep features \cite{Liang2024FusionFD} to optimize the model. Nevertheless, most existing methods fail to fully account for the underlying mechanisms of different tasks,  limiting their adaptability and interpretability \cite{Zhu2024TaskCustomizedMO}. 

\begin{figure}[!t]
	\vspace{-4pt}
	\centering
	\!\!\includegraphics[width=7.8cm]{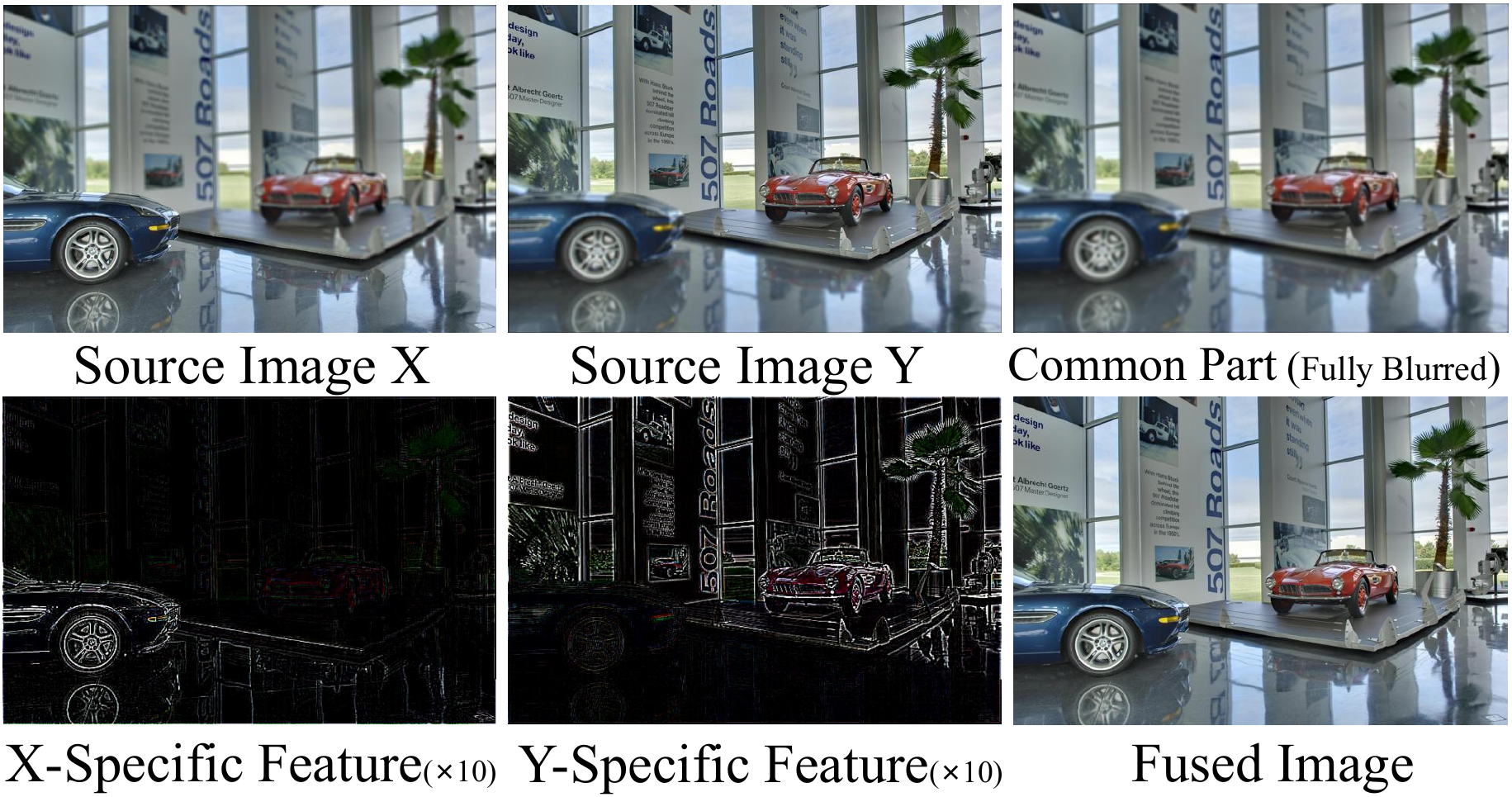}
	
	\vspace{-5pt}
	\caption{Visual illustration of the fusion process for MFF task.}
	\label{f0}
    \vspace{-10pt}
\end{figure}

\vspace{-2pt}
To address these issues, we propose a mechanism-aware unsupervised general image fusion (MAUGIF) method based on Dual Cross-Image  Autoencoders (DCIAE). According to the inherent mechanisms of different fusion tasks, we introduce a classification of additive and multiplicative fusion.  For example, the fused output of VIF can be  modeled as a combination where thermal saliency complements visible details in an approximately additive manner, thus belonging to additive fusion. In contrast, HMF involves spatial  deblurring and super-resolution of  HSIs, which can be  modeled through multiplicative interactions rather than direct addition.
Then, dual encoders are employed to extract the common content from source images, while modality-specific details are isolated. The dual decoders then act as feature injectors, reconstructing images by reintegrating their unique characteristics into the shared content. In the fusion process,  the modality-specific features are injected into the source image, thereby generating the fused image that integrates  information from both modalities. Notably, different fusion mechanisms correspond to distinct decoder architectures, i.e., different feature injection manners, enhancing both the performance and interpretability. Moreover, our model  consists of two lightweight encoder–decoder branches, ensuring a compact architecture. The fusion process  is accomplished in a single decoding step, which eliminates the need for designing complex fusion rules. As demonstrated in Figure \ref{f11}, this design leads to high computational efficiency. This makes our method well-suited for real-time applications and deployment in resource-constrained environments.
We summarize the main contributions of this paper as follows.
\vspace{-2pt}
\begin{itemize}
    \item \textbf{An interpretable and efficient general image fusion framework.} We propose a mechanism-aware general image fusion method termed MAUGIF, which explicitly disentangles the shared content and modality-specific details of the source images. As shown in Figure \ref{f0},  MAUGIF enables direct visualization  of the fusion process, and  the contribution of each source image  can be clearly observed. In addition, our model is lightweight and the fused image is generated in a single decoding step, leading to high computational efficiency. 

    \item \textbf{A mechanism-aware taxonomy of fusion tasks.} By proposing a classification of fusion into additive and multiplicative types, we can gain insight into how the modality-specific features are integrated. This not only guides the architectural design of the network, but also  enhances the model’s generalization and interpretability.

    \item \textbf{Comprehensive validation across diverse fusion tasks.} Extensive numerical experiments are conducted on various fusion tasks (HMF, VIF, MFF and MEF) to validate the effectiveness and generalization capability of the proposed general image fusion method.
\end{itemize}

\begin{figure*}[t]
    \vspace{-10pt}
    \centering
    \includegraphics[width=0.75\textwidth]{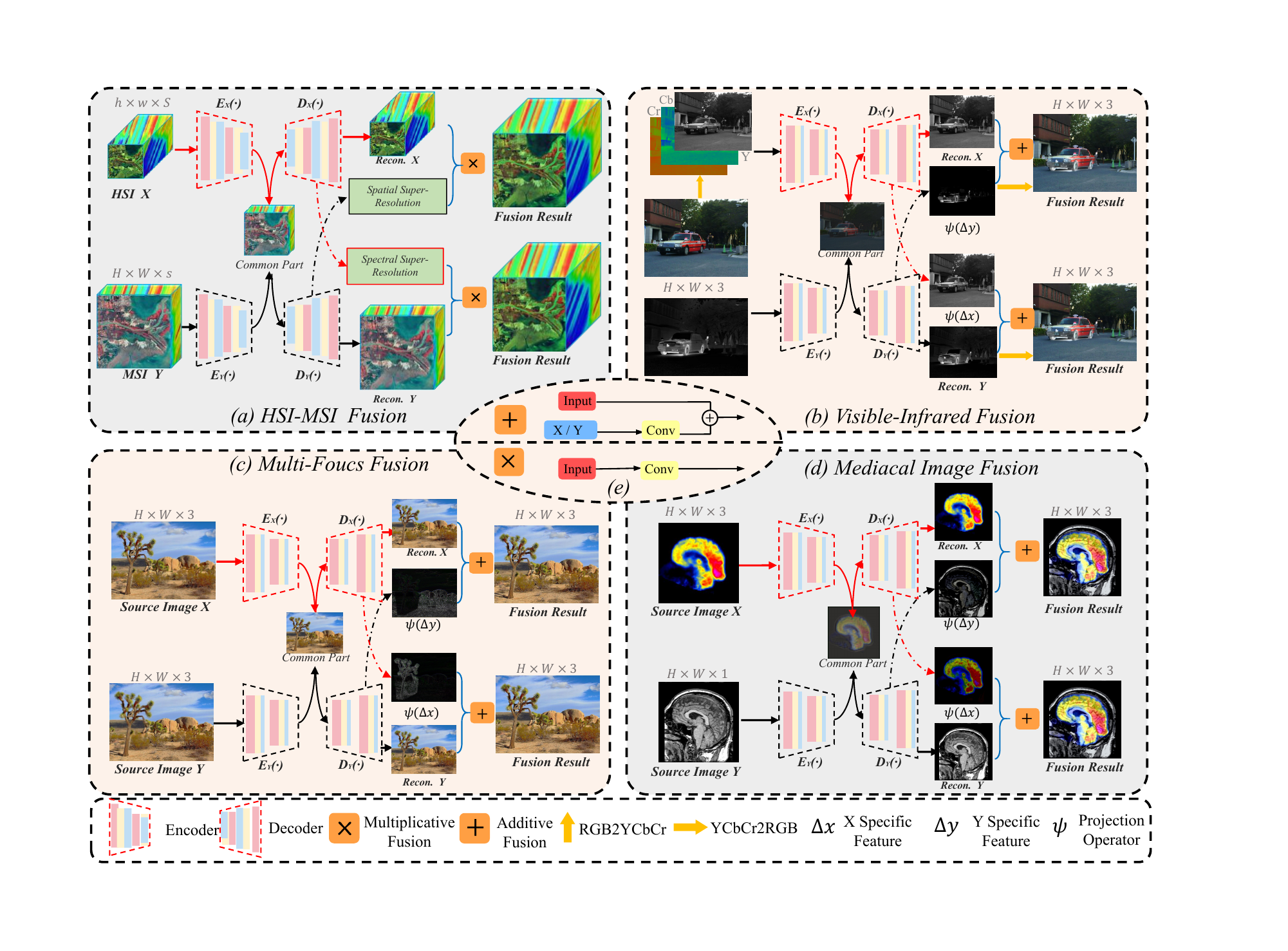}
    \caption{The mechanism-aware DCIAE framework for general image fusion, illustrated with four scenarios.}
    \label{f1}
    \vspace{-0.35cm}
\end{figure*}
\section{Related work}
\vspace{-1pt}
\subsection{Task-specific image fusion methods}
\vspace{-3pt}
Task-specific fusion methods are designed for a particular fusion task, leveraging the inherent properties of input images and prior knowledge to enhance performance. Model-based fusion methods typically rely on mathematical modeling and optimization to exploit image priors \cite{NSSR, OTD}. For example, these methods often leverage various matrix or tensor decompositions to capture and exploit the intrinsic structure of image data \cite{CP1, CSTF}. In addition, they can incorporate image priors using handcrafted regularizers, such as low-rankness \cite{Hysure}, sparsity \cite{WANG2014477}, smoothness \cite{Ma2016}, and nonlocal self-similarity \cite{LTMR}. 
Deep learning-based methods capture high-level semantic content by learning hierarchical representations directly from images. Numerous methods employ convolutional neural networks (CNN) to extract such features from source images \cite{ZSL, Li2019}. Subsequently, generative adversarial networks (GANs) \cite{Ma2019FusionGANAG, Zhang2021a} and vision Transformers \cite{Shao2024, Liu2024} have been introduced to better preserve structural details and global contextual information. Diffusion-based fusion methods have also attracted significant attention \cite{Yue2023, Wu2025}. Furthermore, high-level vision tasks including object detection and semantic segmentation, have been incorporated to guide the fusion process \cite{Tang2022ImageFI, Wang2025DiFusionSegDS}. These methods often deliver high performance on specific fusion tasks but exhibit limited generalizability.

\vspace{-1pt}
\subsection{General image fusion methods}
\vspace{-3pt}
General image fusion  aims to handle multiple fusion tasks within a  unified framework \cite{Ma2022SwinFusionCL}. Recently, Zhang \textit{et al.} \cite{Zhang2020a} proposed a supervised training framework for MFF task, which can be generalized to other tasks by adjusting fusion conditions.   
Ma \textit{et al.} \cite{Ma2022} employed a unified network architecture and loss function, but trained separate models for each individual task. Wang \textit{et al.} \cite{Wang2024a} proposed a  semi-supervised learning approach, which decomposes the source images into features of different frequencies and adaptively  merges the  features. The work in \cite{Zhu2024} accommodated diverse fusion tasks by dynamically customizing a mixture of task-specific adapters,  preserving task individuality while leveraging cross-task commonalities.

\section{The proposed method}
In this paper, the source images are denoted as \( \textbf{X} \) and \( \textbf{Y} \), respectively, and the fused image is denoted as \( \textbf{F} \).

\subsection{Mechanism-Aware Image Fusion}
To capture the diverse fusion mechanisms of information integration, we classify image fusion into additive and multiplicative types. The classification depends on whether  modality-specific features are added to the target or modulate it.
The specific definitions are as follows.
\begin{defn}(\textbf{Additive fusion})
	Given source images \textbf{X} and \textbf{Y} capturing the same scene, \textit{additive fusion} aims to generate a composite image \textbf{F} such that:
	\begin{equation}\label{C7}
        \vspace{-0.1cm}
		\textbf{F}=\textbf{Y}+\psi(\Delta_x) \quad\text{or} \quad \textbf{F}=\textbf{X}+\psi(\Delta_y),
	\end{equation}
	where $\psi$ is a non-expansive projection, i.e., $\Vert\psi(x)-\psi(y)\Vert\le \Vert x-y \Vert$,  $\Delta_x:=\textbf{X}-\textbf{C}$ and $\Delta_y:=\textbf{Y}-\textbf{C}$ are  modality-specific features unique to \textbf{X} and \textbf{Y}, respectively, and $\textbf{C}$ is the common content of the same size as \textbf{X} and \textbf{Y}.
\end{defn}

\vspace{-0.1cm}
For example, in visible-infrared image fusion, the goal is to highlight the thermal radiation of the target from the infrared image (IR) while preserving the structural content of the visible image, which is typically considered additive fusion. However, IR image often contains redundant regions and noise,  which should not be blindly injected into the fused image. Therefore, $\psi$ acts as a  thresholding mechanism to suppress  irrelevant responses in  modality-specific components. The non-expansiveness of $\psi$ ensures that the features are selectively preserved, but never amplified.

\begin{remark}(The choice of $\psi$)
	The design of $\psi$ is adapted to the specific fusion task. For MFF where the modality-specific information of images is all needed, $\psi$ can be set to the identity mapping, i.e., $\psi(x)=x$,  to preserve all fine details. For VIF, a hard-thresholding:
	\begin{equation}\label{C77}
		\psi(x)= x \quad \text{if} ~~ x>\sigma; \quad \text{otherwise}~~ 0,
        \vspace{-0.1cm}
	\end{equation}
	can be employed to suppress weak or spurious responses and enhance salient thermal  targets, where $\sigma\in [0, ~0.4] $.
\end{remark}

\begin{defn}(\textbf{Multiplicative fusion}) Given source images \textbf{X} and \textbf{Y} capturing the same scene,
	\textit{multiplicative fusion} generates the fused image \textbf{F} via modulation like:
    \vspace{-0.1cm}
	\begin{equation}\label{C8}
		\textbf{X}=\textbf{R}*\textbf{F} \quad \text{and}\quad \textbf{Y}=\textbf{F}*\textbf{B},
        \vspace{-0.1cm}
	\end{equation}
	where \textbf{R} and \textbf{B} are degradation fields, `*' denotes matrix multiplication or Hadamard product, etc. Equation (\ref{C8}) indicates that \textbf{X} and \textbf{Y} are different degraded versions of \textbf{F}.
\end{defn}

For instance, hyperspectral-multispectral image fusion (HMF) model is typically formulated as:
\begin{equation*}
\vspace{-0.1cm}
    \min_{\textbf{F}} \Vert \textbf{FB}-\textbf{X} \Vert_F^2+\Vert \textbf{RF}-\textbf{Y} \Vert_F^2,
    \vspace{-0.1cm}
\end{equation*}
where \textbf{X} and \textbf{Y} denote the HSI and MSI, \textbf{B} and \textbf{R} represent the spatial and spectral degradation operators, respectively \cite{Hysure}. Clearly, HMF belongs to multiplicative fusion.

\subsection{Dual cross-image autoencoders framework}
To achieve general image fusion, it is crucial to identify  the fundamental commonalities shared across diverse image fusion tasks. We observe that:
\begin{itemize}
    \item Due to modality differences, images \textbf{X} and \textbf{Y} are inherently distinct, each with unique features, i.e., $\textbf{X}\neq\textbf{Y}$.

	\item  Despite modality differences, $\mathbf{X}$ and $\mathbf{Y}$ correspond to the same scene and thus share common information, which we denote as $\mathcal{C}(\mathbf{X}, \mathbf{Y}) \neq \emptyset$, where $\mathcal{C}(\cdot,\cdot)$ represents the set of shared semantic or structural components.

\end{itemize}
Based on this, we propose the dual cross-image autoencoder (DCIAE) framework for general image fusion, as displayed in Figure \ref{f1}. The  process can be divided into three steps:

\textbf{1)} To extract the common components of source images, we employ two encoders, denoted as \( E_x(\cdot) \) and \( E_y(\cdot) \), to  encode  \( \textbf{X} \) and \( \textbf{Y} \), respectively, such that $
E_x(\textbf{X}) \approx E_y(\textbf{Y})
$.
The loss function for the encoders can be formulated as:
\begin{equation}\label{C1}
	loss_e = loss_1+\lambda loss_2,
\end{equation}
where $\lambda>0$ is a weight parameter, 
\begin{equation}
\vspace{-3pt}
	loss_1 := \Vert E_x(\textbf{X})-E_y(\textbf{Y}) \Vert_F^2,
\end{equation}
\begin{equation}\label{l2}
	loss_2 := \Vert E_x(\textbf{X})-\textbf{X} \Vert_F^2+\Vert \text{E}_y(\textbf{Y})-\textbf{Y} \Vert_F^2.
\end{equation}
Here, $loss_1$ encourages the encoders to extract the shared content between images $\mathbf{X}$ and $\mathbf{Y}$, while $loss_2$ prevents degenerate solutions like $E_x(\mathbf{X}) = E_y(\mathbf{Y}) = 0$.  Notably, $loss_2$ ensures that the learned common components remain faithful to the source images and enforces it to have the same size as the source images. This aims primarily to extract  common content that aligns with  visual intuition.

\begin{figure}[t]
    \vspace{-3pt}
    \centering
    \includegraphics[width=0.3\textwidth]{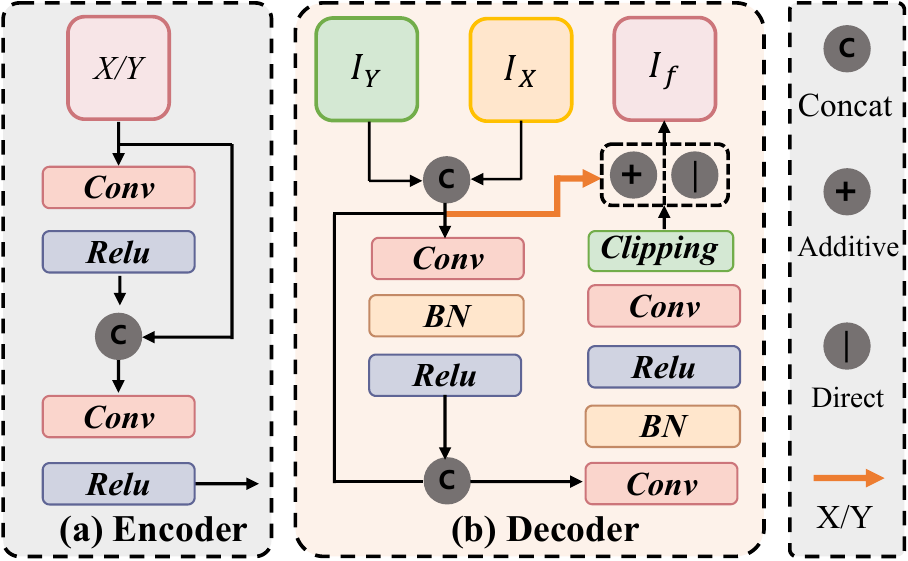}
    \vspace{-0.1cm}
    \caption{Architecture of the Encoder and Decoder.}
    \label{ModelArch}
    \vspace{-0.4cm}
\end{figure}

\textbf{2)} Two decoders, denoted by \( D_x(\cdot) \) and \( D_y(\cdot) \), are employed to decode the learned shared content. The objective is to reconstruct the source images \( \textbf{X} \) and \( \textbf{Y} \) from the common components, i.e.,
\begin{equation}\label{C4}
	D_x(E_x(\textbf{X})) = \textbf{X},\quad D_y(E_y(\textbf{Y})) = \textbf{Y}.
\end{equation}
\textit{The decoder acts as a feature injector, which injects the source image's modality-specific features into the shared content, thereby reconstructing the source images.} The loss function for the decoders can be defined as:
\begin{equation}
	loss_d:= \Vert D_x(E_x(\textbf{X}))-\textbf{X}\Vert_F^2+\Vert D_y(E_y(\textbf{Y}))-\textbf{Y}\Vert_F^2.
\end{equation}
Their architecture  is shown in Figure \ref{ModelArch}.

\textbf{3)} Finally, we perform image fusion by injecting the modality-specific features  of image \( \textbf{X} \) into image \( \textbf{Y} \), or vice versa. Therefore, the  fused image \( \textbf{F} \) can be formulated as:
\begin{equation}\label{C6}
	\textbf{F}=D_x(\textbf{Y}) \quad \text{or}\quad \textbf{F}=D_y(\textbf{X}).
\end{equation}
In this process, the decoder serves directly as a fusion operator, shifting the fusion process from conventional rule-based design to the injection of modality specific feature.  

Additive fusion and multiplicative fusion correspond to different decoder architectures, as shown in Figure \ref{f1} (e).  In additive fusion, the decoder architecture follows a residual  paradigm, where the modality-specific feature of \textbf{X} is injected as an additive manner, i.e.,
\begin{equation}\label{C9}
    \vspace{-0.05cm}
	D_x(\cdot)=(\cdot)+\psi(\Delta_x).
\end{equation}
In multiplicative fusion, the decoder instead generates a  modulation field  to refine the structure of the input image.
\begin{remark}
	In DCIAE framework, image fusion is no longer a `black-box' process. We can clearly observe the modality-specific features contributed by the source images during the fusion process.
	Moreover, DCIAE framework is few shot, requiring only one image pair for model training. In addition, the fusion process involves just a single decoding step, making it highly efficient.
\end{remark}

Next, we analyze the factors affecting DCIAE's performance.
Taking additive fusion as an example, we denote
\begin{itemize}
    \item $\mathcal{C}^*$: the ground truth  shared content between \textbf{X} and \textbf{Y}.
    
    \item \( \widehat{\textbf{F}} \): the fusion result of DCIAE under ideal decoder, i.e., \( D_x(E_x(\textbf{X})) = \textbf{X} \) holds exactly.

    \item $\textbf{F}^*$: The optimal fusion result under DCIAE framework.
\end{itemize}
Assume that \(\psi\) is a truncation function selected as needed.
 Then, by combing (\ref{C6}) with (\ref{C9}), one has:
\begin{equation*}\label{D2}
			\begin{aligned}
				\textbf{F} &= \textbf{Y}+\psi(D_x(E_x(\textbf{X}))-E_x(\textbf{X})),\\
				\widehat{\textbf{F}} &= \textbf{Y}+\psi(\textbf{X}-E_x(\textbf{X})),\quad
        \textbf{F}^*=\textbf{Y}+\psi(\textbf{X}-\mathcal{C}^* ).
			\end{aligned}			
		\end{equation*}
According to the non-expansiveness of \( \psi \), one has
		\begin{equation*}\label{D1}
            \begin{aligned}
                \Vert \textbf{F}-\textbf{F}^* \Vert&\le \Vert \textbf{F}-\widehat{\textbf{F}} \Vert+\Vert \widehat{\textbf{F}}-\textbf{F}^* \Vert\\
                &\le  \Vert D_x(E_x(\textbf{X}))-\textbf{X} \Vert+\Vert  E_x(\textbf{X})-\mathcal{C}^* \Vert.
            \end{aligned}
		\end{equation*}
This indicates that the performance of the DCIAE framework is mainly influenced by two factors: the network's expressive capacity and the accuracy of the shared content.

\section{Experiments}
In this section, we validate the effectiveness of the proposed MAUGIF method on four tasks, including hyperspectral-multispectral image fusion (HMF), visible-infrared image fusion (VIF), multifocus image fusion (MFF), and medical image fusion (MEF).  We also conduct a comparison of computational complexity and ablation experiments.


\subsection{Experimental setup}
\subsubsection{Implementation details}
All experiments are conducted on Ubuntu 20.04 with PyTorch 2.2.0. The hardware configuration includes an Intel Core i7-13700KF CPU with a clock speed of 3.4GHz, 32GB of RAM, and an NVIDIA GeForce RTX 4090 GPU with 24GB of VRAM. Throughout the training process, the AdamW \cite{Loshchilov2017DecoupledWD} optimizer is employed, along with a linear learning rate schedule that incorporates a warm-up phase. The maximum learning rate is set to $1\times 10^{-4}$, the batch size is set to 32, and the training runs for 50 epochs.

\subsubsection{Dataset}
Next, we introduce the datasets used in our experiments.

\textbf{CAVE$^1$ and Harvard Dataset \cite{data11}  for HMF}: For the CAVE dataset, we selected  `Toys' and `Peppers', whose spatial dimensions  were downsampled by different scale factors (\textbf{sf}). We simulated  MSIs with three bands by the Nikon D700 camera$^2$.
For the Harvard dataset,  images `img3' and `imgc6'   are selected. Each  contains a 31-band HSI, which serves as the ground truth. The low-resolution HSIs are generated through downsampling the original HSI at different scale factors. We divide the  spectral bands of HSI into three parts and generate a 3-band MSI by averaging the bands. We corrupt  HSI and MSI with  Gaussian noise to obtain SNRs of 35 dB and 40 dB, respectively.

\textbf{MSRS Dataset \cite{Tang2022ImageFI} for VIF:} This dataset is a benchmark for the  fusion and segmentation of infrared-visible images. It comprises 1,444 well-registered image pairs. Following the official protocol, we adopt the original split with 1,083 pairs for training and 361 pairs for testing. All images have a resolution of 640×480 and are accompanied by pixel-level semantic segmentation annotations.

\textbf{MFI-WHU Dataset \cite{zhang2021mff} for MFF:} This dataset is a benchmark for multi-focus image fusion, which contains 120 pairs of  images, synthesized with Gaussian blurring and provided with manually annotated decision maps.

\textbf{Harvard Dataset$^3$ for MEF:} This dataset is a well-known medical imaging benchmark for multimodal fusion. It provides a large number of CT–MRI, PET–MRI, and SPECT–MRI image pairs, which are widely used to train and evaluate medical image fusion models.

\footnotetext[1]{https://www.cs.columbia.edu/CAVE/databases/multispectral/ }
\footnotetext[2]{https://maxmax.com/spectral\_response.htm}
\footnotetext[3]{http://www.med.harvard.edu\/AANLIB/home.html}

\begin{figure*}\label{H1}
\vspace{-10pt}
	\centering
	\begin{subfigure}{0.116\linewidth}  
		\includegraphics[width=\linewidth]{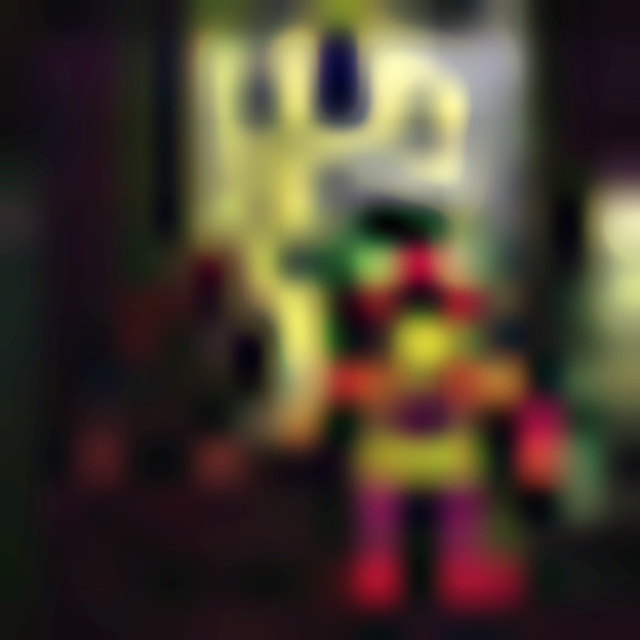}
	\end{subfigure}
	\begin{subfigure}{0.116\linewidth}
		\includegraphics[width=\linewidth]{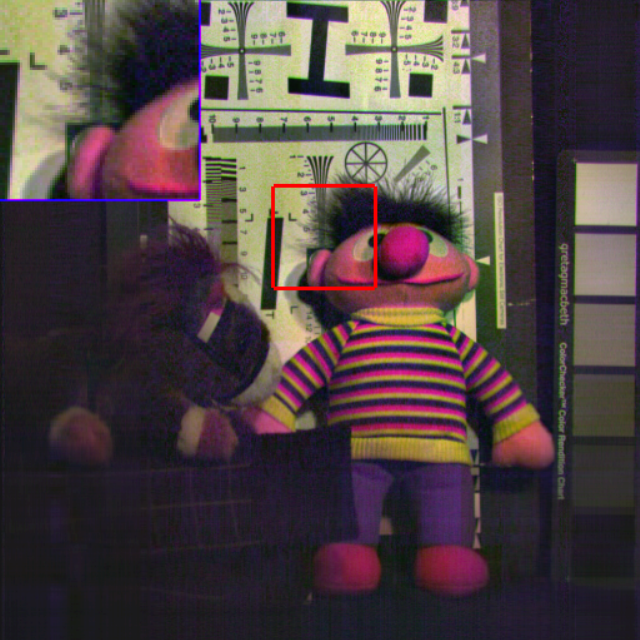}
	\end{subfigure}
	\begin{subfigure}{0.116\linewidth}
		\includegraphics[width=\linewidth]{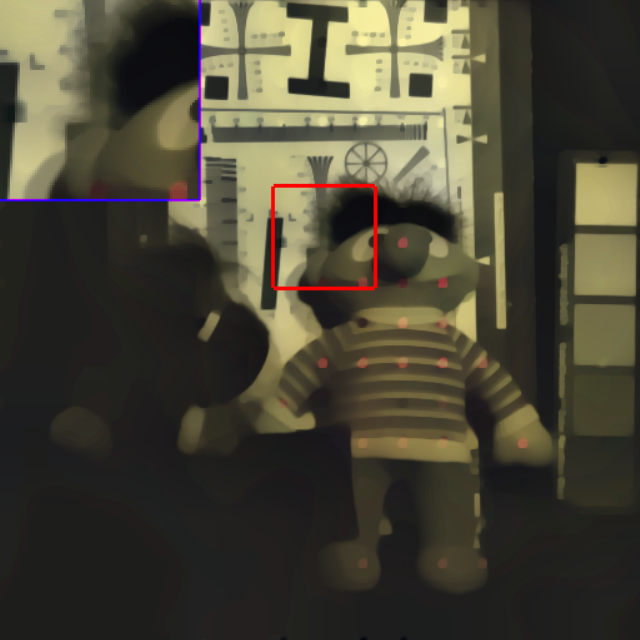}
	\end{subfigure}
	\begin{subfigure}{0.116\linewidth}
		\includegraphics[width=\linewidth]{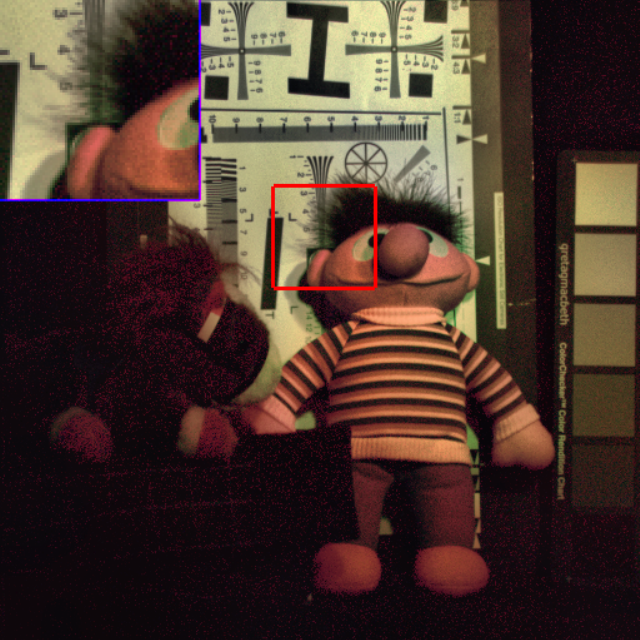}
	\end{subfigure}
	\begin{subfigure}{0.116\linewidth}
		\includegraphics[width=\linewidth]{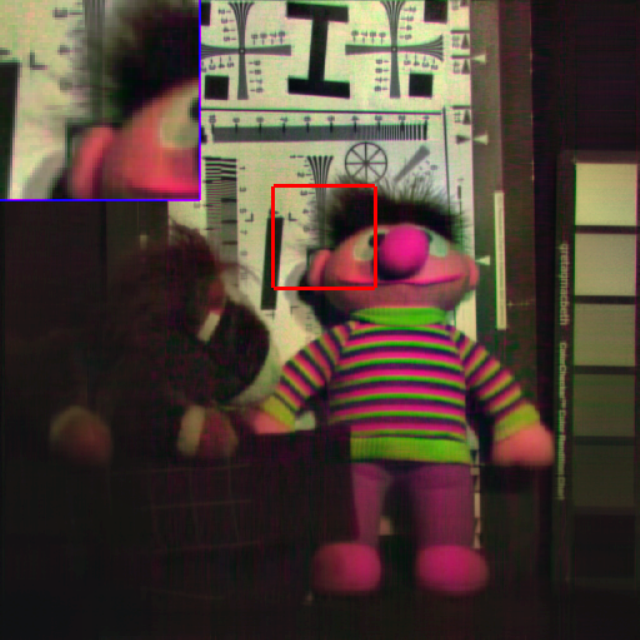}
	\end{subfigure}
	\begin{subfigure}{0.116\linewidth}
		\includegraphics[width=\linewidth]{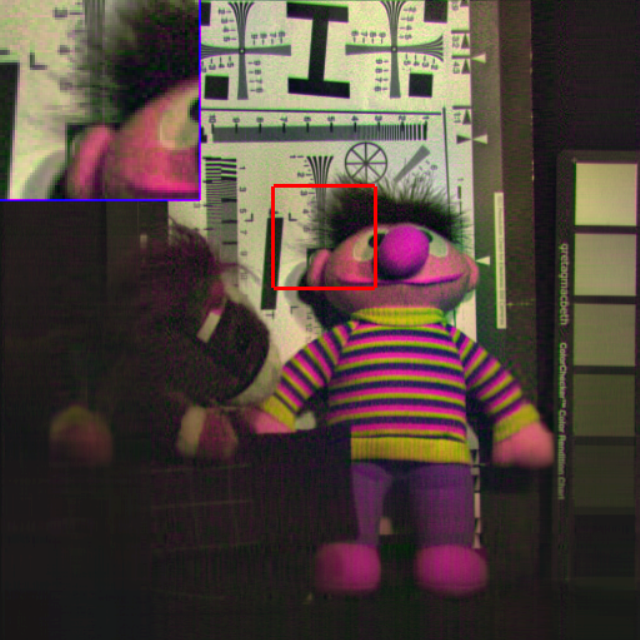}
	\end{subfigure}
	\begin{subfigure}{0.116\linewidth}
		\includegraphics[width=\linewidth]{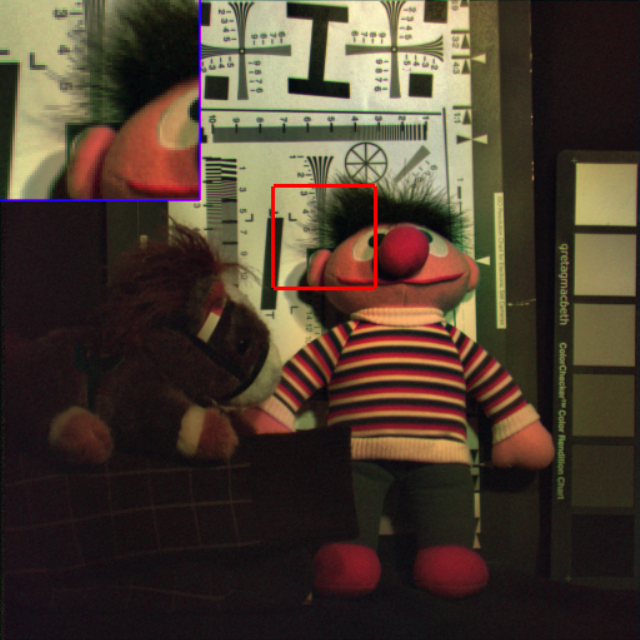}
	\end{subfigure}
	\begin{subfigure}{0.116\linewidth}
		\includegraphics[width=\linewidth]{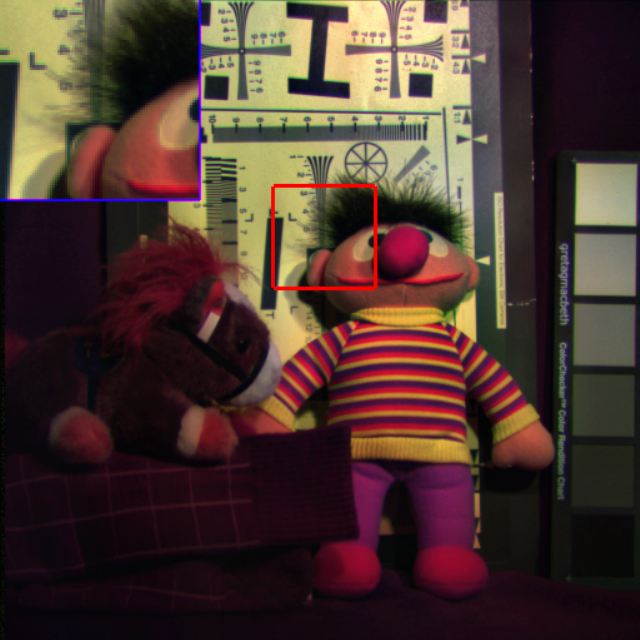}
	\end{subfigure}
	
	\begin{subfigure}{0.116\linewidth}
		\includegraphics[width=\linewidth]{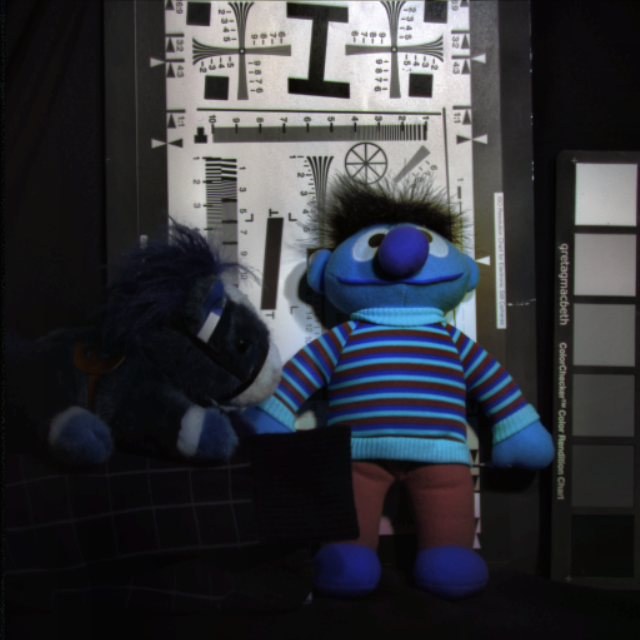}
		\caption{HSI/MSI}
	\end{subfigure}
	\begin{subfigure}{0.116\linewidth}
		\includegraphics[width=\linewidth]{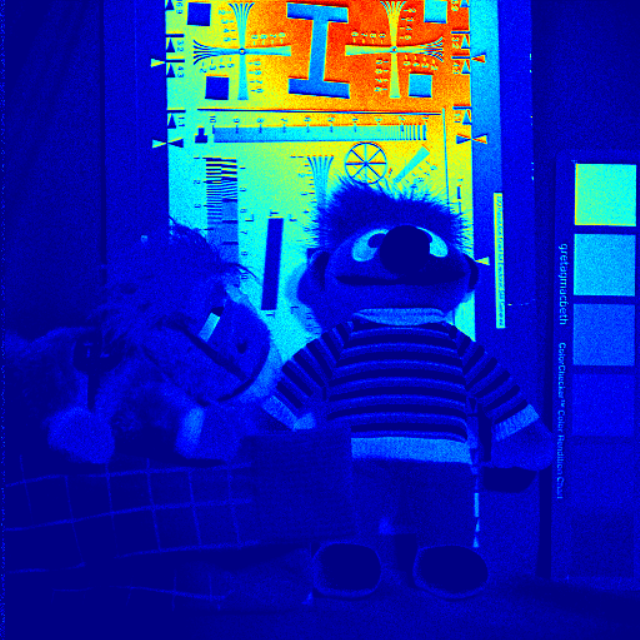}
		\caption{CNMF}
	\end{subfigure}
	\begin{subfigure}{0.116\linewidth}
		\includegraphics[width=\linewidth]{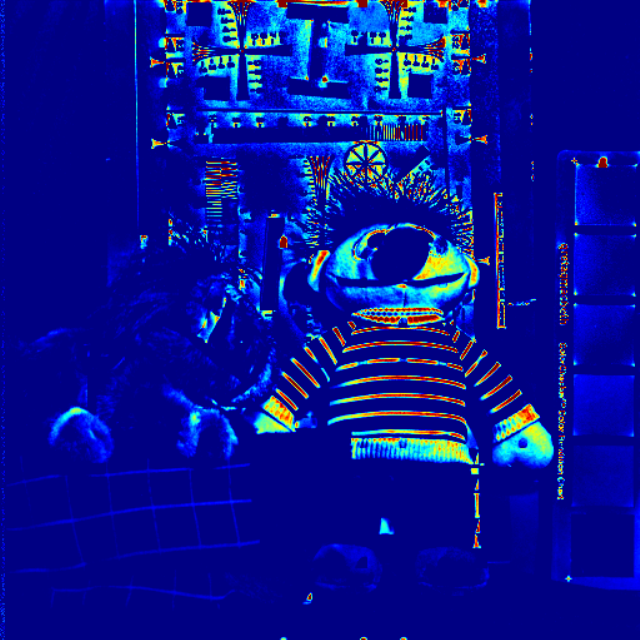}
		\caption{Hysure}
	\end{subfigure}
	\begin{subfigure}{0.116\linewidth}
		\includegraphics[width=\linewidth]{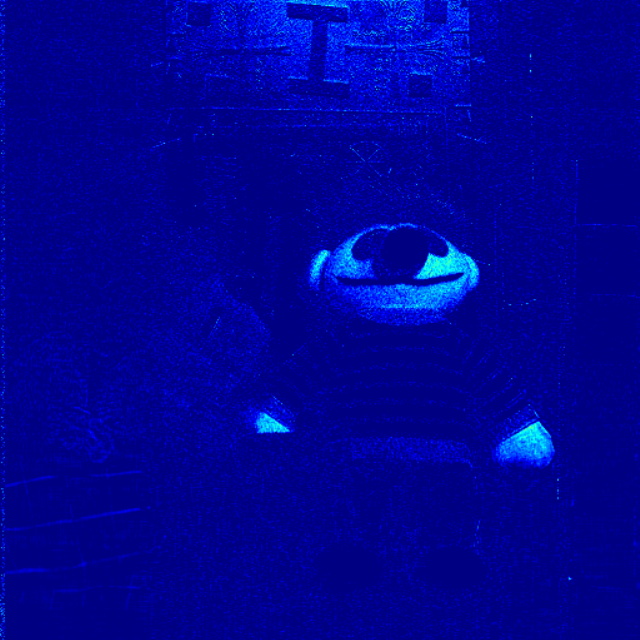}
		\caption{ZSL}
	\end{subfigure}
	\begin{subfigure}{0.116\linewidth}
		\includegraphics[width=\linewidth]{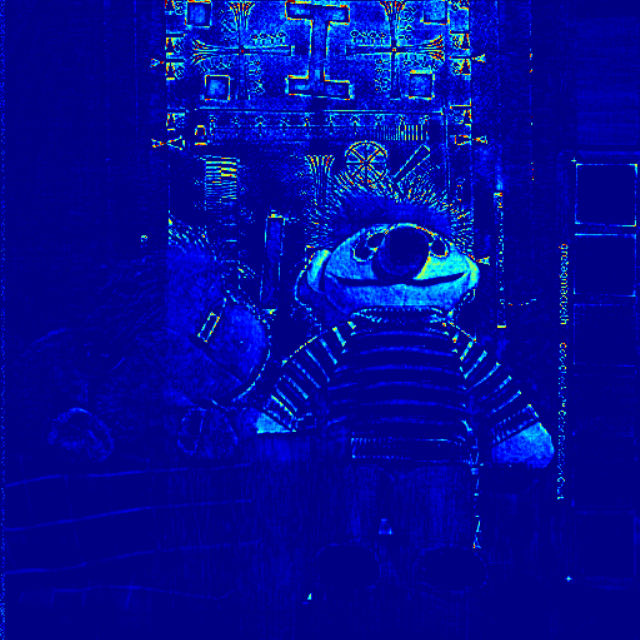}
		\caption{TriDLR}
	\end{subfigure}
	\begin{subfigure}{0.116\linewidth}
		\includegraphics[width=\linewidth]{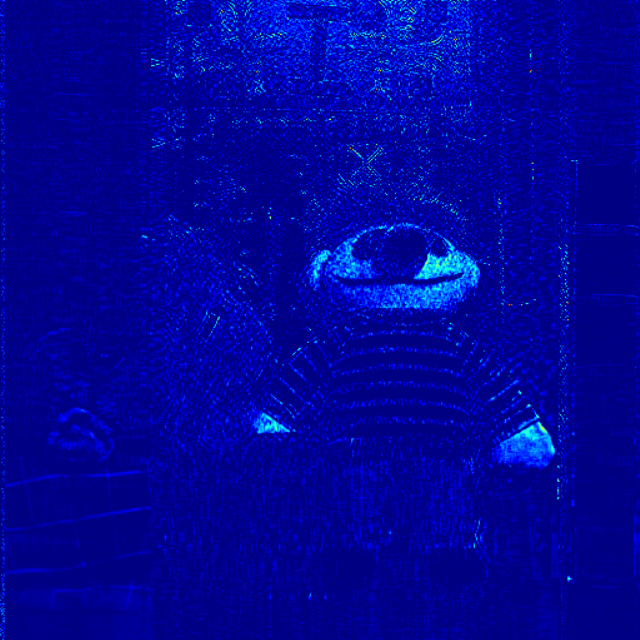}
		\caption{DTDNML}
	\end{subfigure}
	\begin{subfigure}{0.116\linewidth}
		\includegraphics[width=\linewidth]{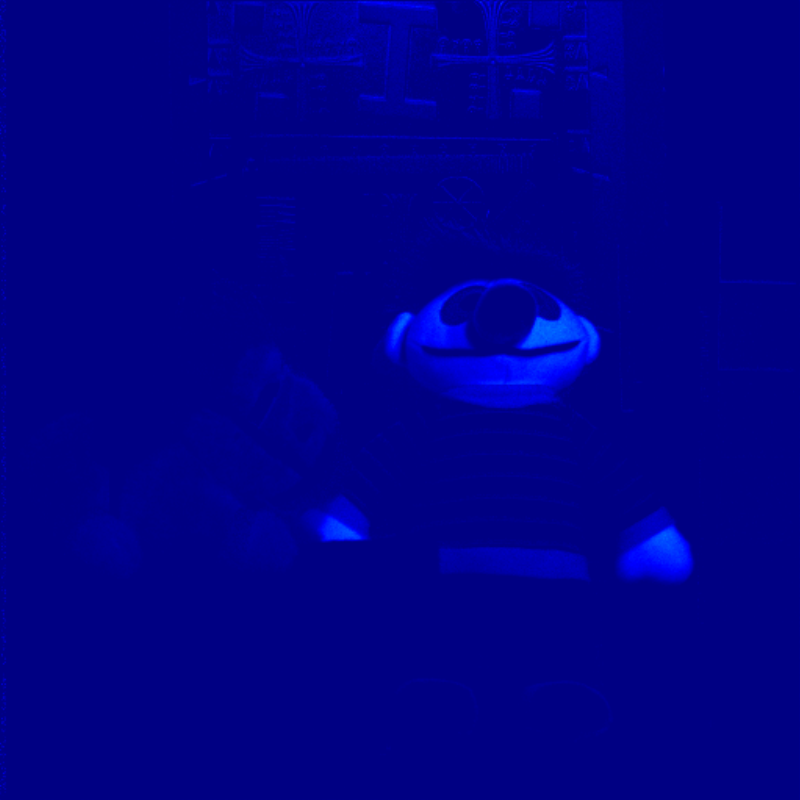}
		\caption{MAUGIF}
	\end{subfigure}
	\begin{subfigure}{0.116\linewidth}
		\includegraphics[width=\linewidth]{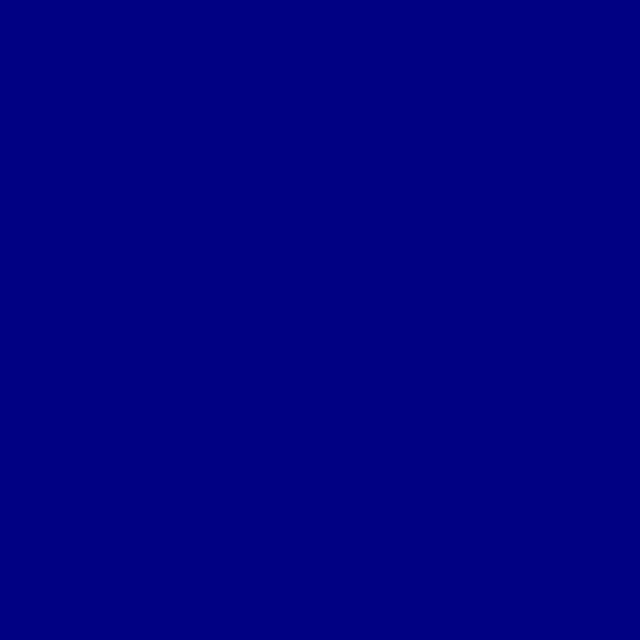}
		\caption{GT}
	\end{subfigure}

    \begin{subfigure}{0.4\linewidth}
		\includegraphics[width=\linewidth]{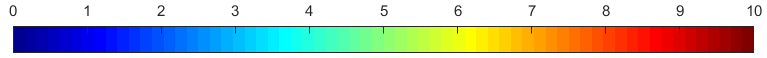}
		\caption*{}
	\end{subfigure}

    \vspace{-0.6cm}
	\caption{The fused images of compared methods and the corresponding error maps for `Toys' dataset. \label{H2}}
\end{figure*}
\begin{figure*}[t]
    \centering
    \includegraphics[width=0.96\textwidth]{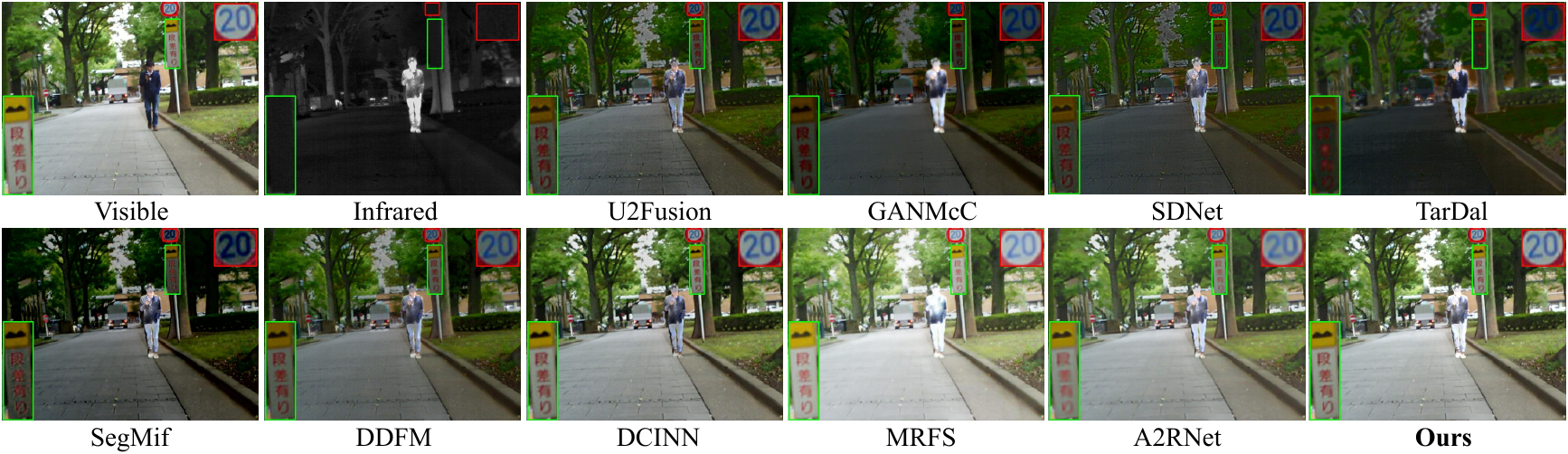}

    \vspace{-0.15cm}
    \caption{Qualitative comparison of MAUGIF with 9 SOTA methods on image 00500D from the MSRS visible-infrared dataset.}
    \label{VifFusion}
    \vspace{-0.2cm}
\end{figure*}

\subsection{Comparison on HMF task}
In hyperspectral-multispectral image fusion (HMF), the  MAUGIF approach performs fusion without relying on the knowledge of the degradation process, and thus falls into the category of blind fusion \cite{CP1}. Therefore,  we select five blind HMF methods for comparison, including the classical CNMF \cite{CNMF} and HySure \cite{Hysure}, as well as the recently proposed ZSL \cite{ZSL}, TriDLR \cite{Yang2025}, and DTDNML \cite{Wang2024}.
\begin{table}[hpbt]\scriptsize
	\renewcommand\arraystretch{1.07}
	\caption{The evaluation metrics of the fused images for HMF task.} 
	\vspace{-0.25cm}      
	\centering
    \resizebox{0.5\textwidth}{!}{
	\begin{tabular}{p{0.7cm}p{0.9cm}p{0.6cm}p{0.6cm}p{0.6cm}p{0.6cm}p{0.6cm}}
		\hline
		Dataset& Method & PSNR & SSIM & ERGAS & SAM& Time(s) \\ \hline
		& CNMF & 32.03 & 0.881 & 2.199 & 17.69 & \textcolor{red}{9.9} \\
		& Hysure & 28.43 & 0.866 & 3.096 & 23.74 & 40.5\\ 
		\textbf{Peppers}& ZSL & \textcolor{blue}{40.32} & \textcolor{blue}{0.978} & \textcolor{blue}{1.172} & \textcolor{blue}{11.95} & 387.1\\ 
		~\textbf{(sf=16)}& TriDLR & 36.85 & 0.952 & 1.353 & 15.19 & 57.3\\  
		& DTDNML & 37.42 & 0.969 & 1.302 & 14.79 & 31.8 \\
		& MAUGIF & \textcolor{red}{40.91} &\textcolor{red}{0.982} & \textcolor{red}{1.038} & \textcolor{red}{11.87} & \textcolor{blue}{10.7} \\ \cline{2-7} 
		& CNMF & 29.75 & 0.872 & 0.647 &16.79 & \textcolor{red}{8.6} \\
		& Hysure & 27.01 & 0.832 & 0.779 & 21.74 & 36.1 \\ 
		~\textbf{Toys}& ZSL & \textcolor{blue}{38.27} & \textcolor{blue}{0.964} & \textcolor{blue}{0.408} & \textcolor{blue}{11.37} & 325.8 \\ 
		\textbf{(sf=32)}& TriDLR & 34.50 & 0.948 & 0.438 &12.43 & 46.9 \\  
		& DTDNML & 34.86 & 0.953 & 0.436 &13.28 & 27.5 \\
		& MAUGIF & \textcolor{red}{39.68} & \textcolor{red}{0.980} & \textcolor{red}{0.297}& \textcolor{red}{9.22} &  \textcolor{blue}{8.9} \\
        \hline
		& CNMF & 38.59 & 0.943 & 3.734 & 11.97 & \textcolor{red}{11.2} \\
		& Hysure & 40.161 & 0.949 & 3.512 & 8.69 & 56.5\\ 
		~\textbf{img3}& ZSL & \textcolor{blue}{42.32} & \textcolor{blue}{0.972} & \textcolor{blue}{2.528} & \textcolor{red}{8.75} & 487.5 \\
		\textbf{(sf=8)}& TriDLR &  41.97 & 0.969 & 2.803 &9.48 & 71.4 \\ 
		& DTDNML & 41.06 & 0.970 & 2.742 & 9.66 & 38.7 \\  
		& MAUGIF & \textcolor{red}{42.90} & \textcolor{red}{0.975} & \textcolor{red}{2.431} & \textcolor{blue}{9.07} & \textcolor{blue}{12.4} \\ \cline{2-7}
		& CNMF & 37.14 & 0.954 & 0.923 & 6.02 &  \textcolor{red}{9.4}\\
		& Hysure & 37.87 & 0.957 & 0.983 & 5.79 & 48.3 \\ 
		~\textbf{imgc6}& ZSL & \textcolor{blue}{39.82} & \textcolor{blue}{0.967} & \textcolor{blue}{0.867} & 5.45 & 420.3 \\ 
		\textbf{(sf=16)}& TriDLR & 38.64 & 0.962 & 0.927 & \textcolor{red}{4.92} & 64.7 \\  
		& DTDNML & 37.76 & 0.958 & 0.948 & 5.85 & 34.9\\
		& MAUGIF & \textcolor{red}{40.76} & \textcolor{red}{0.969} & \textcolor{red}{0.853} & \textcolor{blue}{5.03} & \textcolor{blue}{10.8} \\ 
		\hline
	\end{tabular}}
	\label{Tab1}
	\vspace{-5pt}
\end{table}

Table \ref{Tab1} presents the fusion performance of the compared methods in four HSI datasets. From the table, the CNMF and HySure exhibit relatively weaker performance, particularly in large scale factor cases. In these cases, the fusion model becomes severely ill-posed, which limits their performance. TriDLR reduces the number of parameters through tensor triple decomposition. It adopts a strategy of estimating the degradation operator and performing fusion simultaneously, effectively avoiding error accumulation and achieving superior performance. The ZSL achieves a relatively accurate estimation of the degradation operators, which results in less error accumulation during its subsequent fusion process and leads to competitive fusion performance. However, when the scale factor (\textbf{sf}) is large, the accuracy of the degradation operator estimated by ZSL decreases, leading to a significant drop in fusion performance. In contrast, the proposed MAUGIF method exhibits greater robustness to large \textbf{sf}, as the proposed method can perform spectral super-resolution on the MSI, which is less sensitive to spatial downsampling. Overall, MAUGIF  achieves  optimal performance in four datasets, which demonstrates its effectiveness in HMF task.

\begin{figure*}[htbp]
    \vspace{-10pt}
    \centering
    \!\!\includegraphics[width=0.95\textwidth]{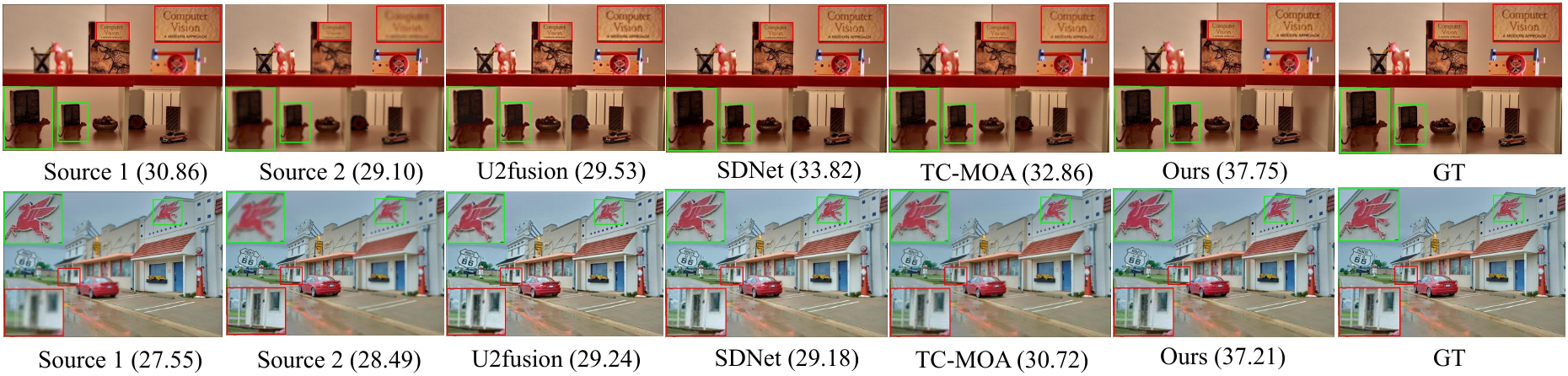}
    \vspace{-5pt}
    \caption{Experimental results of compared methods on three multi-focus image datasets and the corresponding PSNR values.}
    \label{IV}
    \vspace{-0.2cm}
\end{figure*}

\begin{figure*}[htbp]
    \centering
    \begin{subfigure}[b]{0.302\textwidth}
        \includegraphics[width=\textwidth]{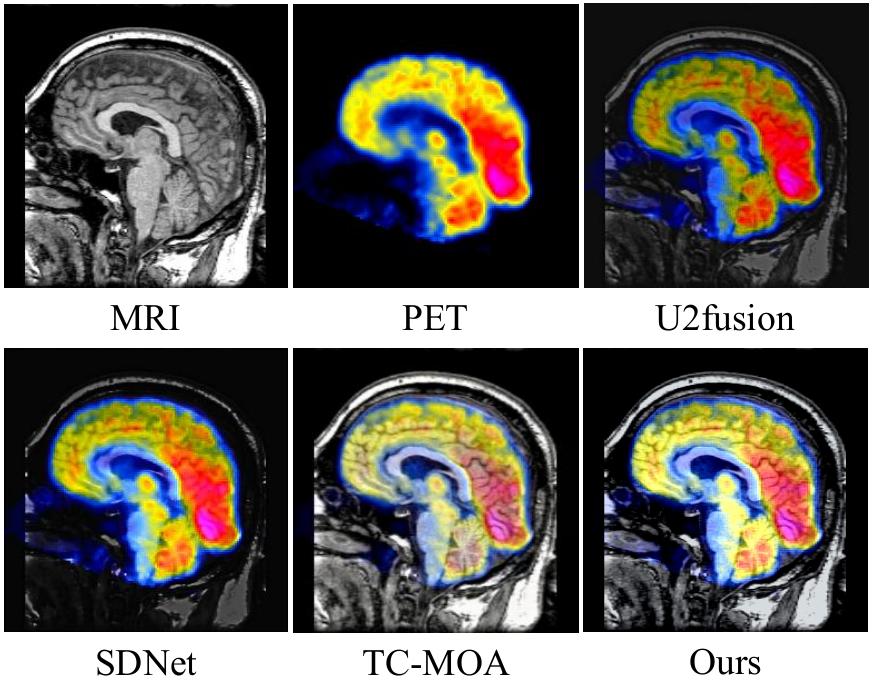}
        \caption{}
        \label{fig:me1}
    \end{subfigure}\quad
    \begin{subfigure}[b]{0.302\textwidth}
        \includegraphics[width=\textwidth]{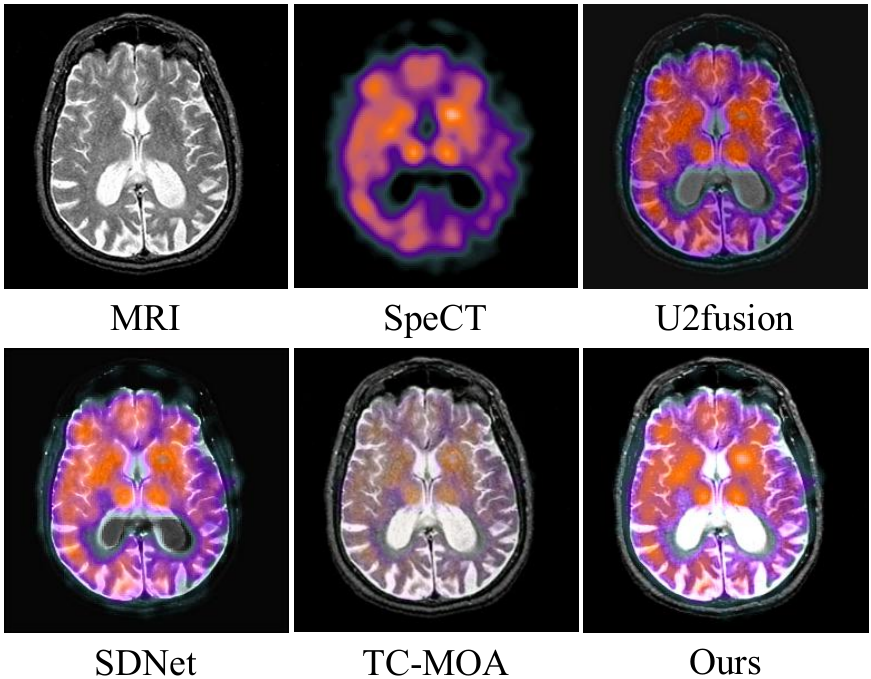}
        \caption{}
        \label{fig:me2}
    \end{subfigure}\quad
    \begin{subfigure}[b]{0.302\textwidth}
        \includegraphics[width=\textwidth]{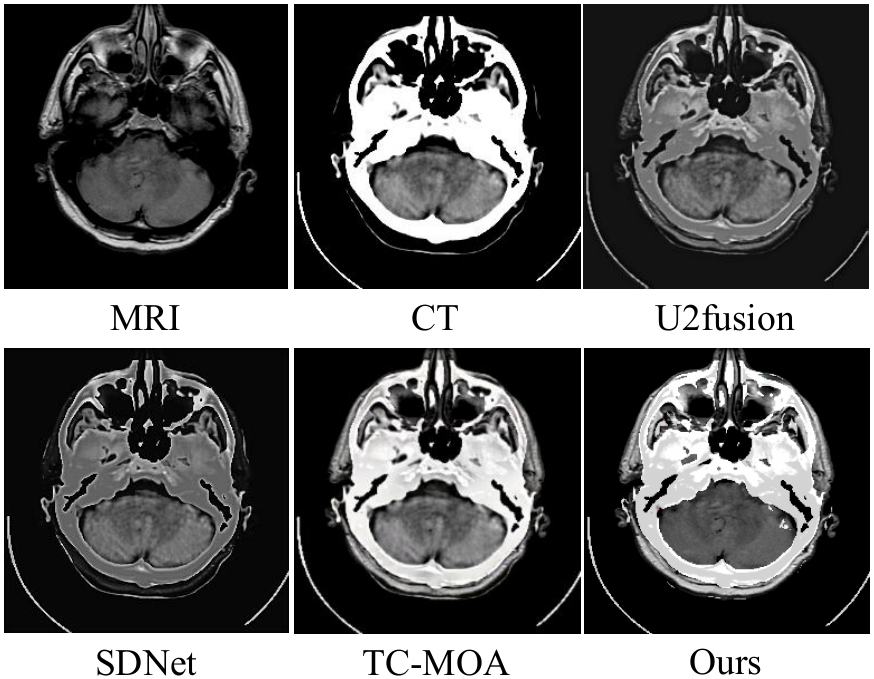}
        \caption{}
        \label{fig:me3}
    \end{subfigure}

    \vspace{-7pt}
    \caption{Experimental results of compared methods on three types of medical image datasets. (a)–(c) correspond to different modalities.}
    \label{V}
    \vspace{-0.5cm}
\end{figure*}

Figure \ref{H2} presents the fused images and error maps  in the `Toys' dataset for visual comparison. 
As shown in the figure, the fused images of  HySure exhibit noticeable color distortions, which indicates a significant estimation error in the spectral degradation operator. As can be seen from the error maps, ZSL achieves competitive results. However,  MAUGIF exhibits the fewest artifacts and  color distortion, achieving the best visual reconstruction quality.

\subsection{Comparison on VIF task}
To evaluate the performance of  MAUGIF on visible-infrared image fusion task, we conduct experiments on  MSRS \cite{Tang2022ImageFI} dataset, and compare the MAUGIF with several state-of-the-art (SOTA) fusion methods. The quantitative results presented in Table \ref{FusionMSRS} demonstrate that MAUGIF achieves five top rankings and one second-place ranking among all competing methods. Specifically, it records an MI of 5.638, an SCD of 1.731, a VIF of 0.949, a \(Q^{AB/F}\) score of 0.652, and an SSIM of 0.482. These results indicate that MAUGIF effectively preserves mutual information, structural similarity, and visual information fidelity, suggesting superior retention of texture, structure, and perceptual quality. 
\begin{table}[htbp]
    \centering
    \caption{ Quantitative comparisons of six metrics on 361 image pairs from the MSRS dataset.}
    \vspace{-0.2cm}
    \label{FusionMSRS}
    \resizebox{0.5\textwidth}{!}{
        \begin{tabular}{lcccccc}
            \toprule
            \textbf{Category} & MI↑ & AG↑ & SCD↑ & VIF↑ & \(Q^{AB/F}\)↑ & SSIM↑  \\
            \midrule
                U2Fusion\textsubscript{20} \cite{Xu2020U2FusionAU}    & 1.988 & 2.515 & 1.145 & 0.516 & 0.386 & 0.381 \\
                GANMcC\textsubscript{21} \cite{Ma2021GANMcCAG}      & 2.459 & 2.188 & 1.458 & 0.612 & 0.326 & 0.309 \\
                SDNet\textsubscript{21} \cite{Zhang2021SDNetAV}      & 1.708 & 2.675 & 0.986 & 0.499 & 0.377 & 0.364 \\
                TarDal\textsubscript{22} \cite{Liu2022TargetawareDA}     & 2.194 & 1.716 & 0.697 & 0.406 & 0.172 & 0.238 \\
                SegMif\textsubscript{22} \cite{Liu2023MultiinteractiveFL}     & 2.472 & 3.336 & \textcolor{blue}{1.594} & 0.774 & 0.565 & 0.339 \\
                DDFM\textsubscript{23} \cite{Zhao2023DDFMDD}       & 2.467 & 2.700   & 1.556 & 0.684 & 0.484 & 0.327 \\
                Dif-Fusion\textsubscript{23} \cite{Yue2023DifFusionTH} & 3.326 & \textcolor{red}{3.889} & 1.592 & \textcolor{blue}{0.827} & \textcolor{blue}{0.583} & \textcolor{blue}{0.448} \\
                DCINN\textsubscript{23} \cite{Wang2023AGP}      & 3.328 & 3.334 & 1.478 & 0.810  & 0.568 & 0.369 \\
                MRFS\textsubscript{24} \cite{Zhang2024MRFSMR}       & 3.068 & 3.034 & 1.288 & 0.74  & 0.484 & 0.350  \\
                A2RNet\textsubscript{25} \cite{li2025a2rnet}     & \textcolor{blue}{3.334} & 2.814 & 1.481 & 0.711 & 0.438 & 0.359 \\
                \textbf{MAUGIF}        & \textcolor{red}{5.638} & \textcolor{blue}{3.457} & \textcolor{red}{1.731} & \textcolor{red}{0.949} & \textcolor{red}{0.652} & \textcolor{red}{0.482}  \\             
            \bottomrule
        \end{tabular}
    }
    \vspace{-0.2cm}
\end{table}
Although Dif-Fusion achieves a slightly higher AG of 3.889, MAUGIF still delivers a competitive AG value of 3.457 and consistently outperforms the other methods across all remaining metrics. These observations validate that MAUGIF can generate fused images with enhanced information content, sharper detail representation, and superior perceptual quality compared with existing SOTA approaches.

To better illustrate the fusion performance of MAUGIF, we provide a qualitative comparison, as shown in Figure \ref{VifFusion}. We can observe that our method preserves fine-grained texture details more effectively than compared approaches. Specifically, in the regions highlighted by the red and green boxes, the text remains sharp, structurally continuous, and clearly recognizable in our results. In contrast, other methods tend to introduce noise interference from the infrared modality, resulting in blurred textures, weakened edge contours, and noticeable degradation of structural fidelity. These visual comparisons further verify the robustness of our fusion strategy in maintaining both clarity and semantic consistency.
In summary, both quantitative and qualitative results demonstrate that MAUGIF preserves complementary information from the source images while maintaining visual fidelity. 

\subsection{The performance on MFF and MEF}
Next, we verify the effectiveness and generalization capability of the proposed MAUGIF model on MFF and MEF tasks. Three general image fusion methods are compared, including U2Fusion \cite{Xu2020U2FusionAU}, SDNet \cite{Zhang2021SDNetAV}, and TC-MOA \cite{Zhu2024TaskCustomizedMO}. The quantitative results are presented in Table \ref{FusionMFF}.
From the table,  the fused images of MFF obtained by the proposed MAUGIF method significantly outperform the compared methods in terms of both PSNR and SSIM metrics. For example, its average PSNR is 4.5 higher than that of the suboptimal SDNet. For MEF, although there is no ground truth image available as a reference, our method achieves the best overall performance in many metrics. This indicates that the fused image preserves both mutual information and visual information fidelity, which fully validates the effectiveness and generalization of our approach.

\begin{table}[htbp]
    \scriptsize
    \centering
    \renewcommand\arraystretch{1.15}
    \caption{Quantitative comparisons  on  image pairs from the MFI-WHU and Harvard  datasets.}
    \label{FusionMFF}
    \resizebox{0.48\textwidth}{!}{
        \begin{tabular}{lcccc}
            \toprule
            \textbf{Method} & U2Fusion\textsubscript{20} & SDNet\textsubscript{21} & TC-MOA\textsubscript{24} & \textbf{MAUGIF} \\
            \midrule
            \multicolumn{1}{c}{} & \multicolumn{4}{c}{\textbf{Comparison on MFF (with GT reference)}} \\
            \cline{2-5}
            PSNR↑ & 28.374 & \textcolor{blue}{32.153} & 31.896 & \textcolor{red}{36.632} \\
            SSIM↑ & 0.917  & \textcolor{blue}{0.955} & 0.949 & \textcolor{red}{0.986} \\ 
            \midrule
            \multicolumn{1}{c}{} & \multicolumn{4}{c}{\textbf{Comparison on MEF (without GT reference)}} \\
            \cline{2-5}
            MI↑  & 2.536 & 2.651 & \textcolor{blue}{2.818} & \textcolor{red}{3.459} \\ 
            AG↑  & 5.088 & 5.265 & \textcolor{blue}{7.501} & \textcolor{red}{7.687} \\ 
            SCD↑ & 0.216 & 0.812 & \textcolor{blue}{1.235} & \textcolor{red}{1.288} \\ 
            VIF↑ & 0.411 & 0.385 & \textcolor{blue}{0.547} & \textcolor{red}{0.563} \\ 
            SF↑  & 17.103 & 23.252 & \textcolor{blue}{24.276} & \textcolor{red}{28.851} \\ 
            \bottomrule
        \end{tabular}
    }
    \vspace{-0.2cm}
\end{table}

Figure \ref{IV} and \ref{V} present visual comparisons of the compared methods on two multi-focus image pairs and three types of medical image datasets. For MFF, our method generates fused images with consistently sharp focus across the entire scene, effectively integrating the in-focus regions from all source inputs.  Moreover, magnified views of the results  reveal that fine details are well preserved with high visual sharpness, whereas other compared methods often suffer from blur or incomplete detail transfer, failing to retain all salient structures effectively.
For MEF task, our method successfully integrates complementary information from different brain imaging modalities. Taking MRI–PET image pairs as an example, U2Fusion and TC-MOA introduce noticeable distortions, while U2Fusion and SDNet fail to preserve fine anatomical structures from the MRI modality, particularly in the red regions, where textures appear blurred or weakened. In contrast, our fused results effectively retain both the structural details of MRI and the metabolic activity patterns of PET, demonstrating superior information preservation. This highlights the effectiveness of MAUGIF in MEF task.

\subsection{Complexity Discussion}
To  evaluate the efficiency of the proposed MAUGIF, we compare it against several SOTA models. We measure performance in terms of floating-point operations (FLOPs), model parameters (Params), and per-frame inference time (Time), as summarized in Table \ref{complexity}.

From the table, MAUGIF exhibits a significant advantage in computational efficiency compared with existing fusion networks. For example, MAUGIF contains only 0.00569M parameters and requires 1.76G FLOPs, which is much lower than other methods. In terms of inference latency, MAUGIF achieves an average processing time of 10.63 ms per image pair, making it the fastest among all compared methods. These results demonstrate that MAUGIF achieves a superior balance between fusion quality and computational cost, making it highly suitable for real-time or resource-constrained applications, such as embedded deployment in intelligent perception systems.
\begin{table}[htbp]
\vspace{-0.2cm}
    \centering
    \caption{\textcolor{black}{Quantitative comparison of different methods in terms of FLOPs, Params, Time on the MSRS dataset.}}
    \vspace{-0.25cm}
    \label{complexity}
    {
    	\renewcommand{\arraystretch}{1.2} 
    	\resizebox{1.0\columnwidth}{!}
    	{
        \begin{tabular}{lrrcc}
            \toprule
            {Method}  & {Input Size} & {Params/M↓} & {FLOPs/G↓} & {Time/ms↓} \\
            \midrule
            U2Fusion\textsubscript{20}\cite{Xu2020U2FusionAU} & 480x640 & 0.66 & 518.00 & 695.64 \\
            SDNet\textsubscript{21}\cite{Zhang2021SDNetAV}  & 480x640  & \textcolor{blue}{0.07} & \textcolor{blue}{64.55} & \textcolor{blue}{122.33} \\
            DDFM\textsubscript{22}\cite{Zhao2023DDFMDD}  & 480x640 & 552.81 & 1.34M & 102760 \\
            Dif-Fusion\textsubscript{23}\cite{Yue2023DifFusionTH}  & 480x640 & 416.47 & 1056.86 & 809.41 \\
            SegMiF\textsubscript{23}\cite{Liu2023MultiinteractiveFL}  & 480x640 & 45.63 & 359.41 & 309.70 \\
            MRFS\textsubscript{24}\cite{Zhang2024MRFSMR}  & 480x640 & 134.96 & 139.13 & 124.72 \\
            TC-MoA\textsubscript{24}\cite{Zhu2024TaskCustomizedMO}  & 480x640 & 340.35 & 3145.68 & 305.35 \\
            DiFusionSeg\textsubscript{25}\cite{Wang2025DiFusionSegDS}  & 480x640 & 40.93 & 174.67 & 77.56 \\
            \textbf{MAUGIF}   & 480x640 &\textcolor{red}{0.00569}  &\textcolor{red}{1.76}  &\textcolor{red}{10.63}  \\
            \bottomrule
        \end{tabular}
    	}
    }
    \vspace{-0.2cm}
\end{table}

\subsection{Ablation studies}
\begin{figure}[htbp]
    \vspace{-0.1cm}
	\centering
	\includegraphics[width=0.48\textwidth]{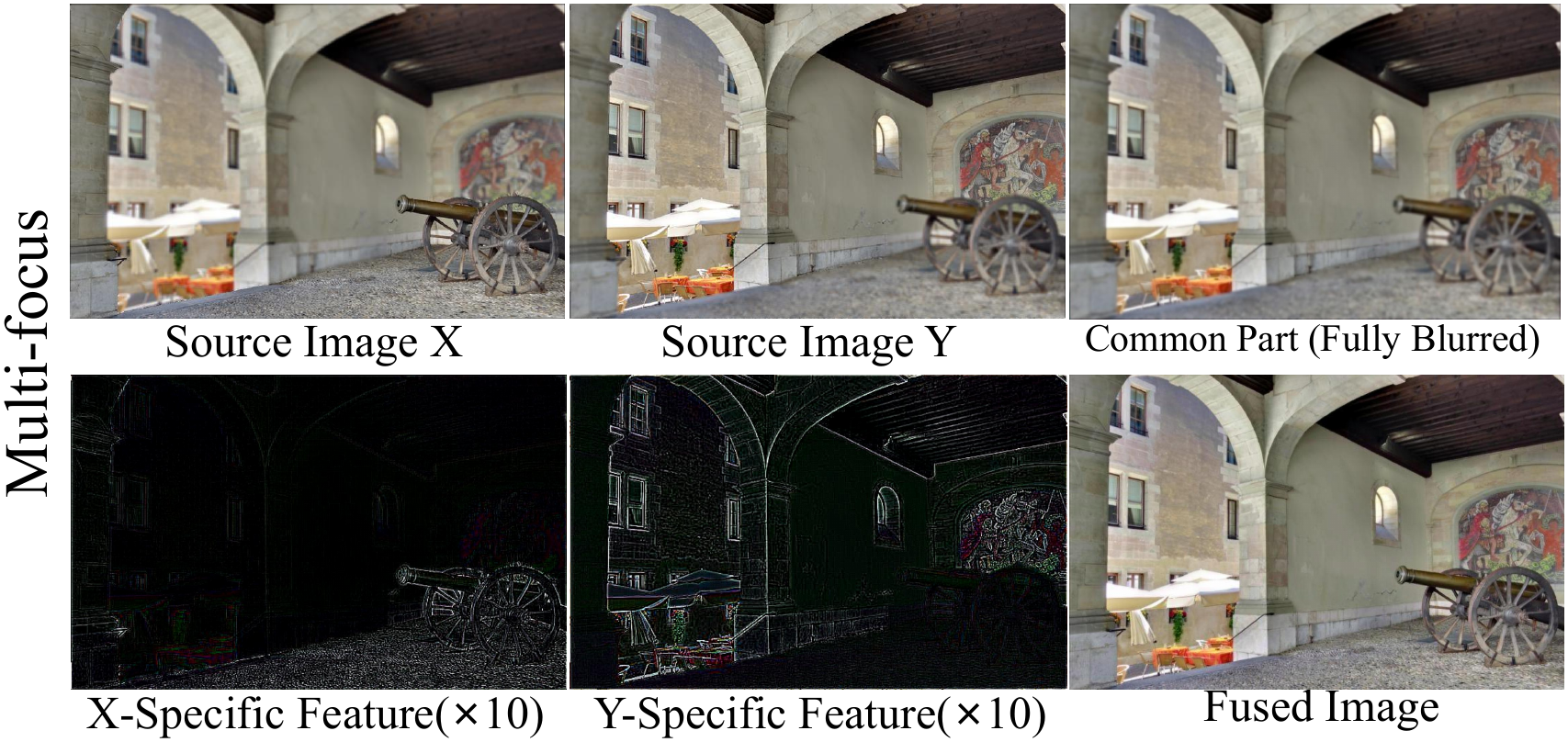}
    \includegraphics[width=0.48\textwidth]{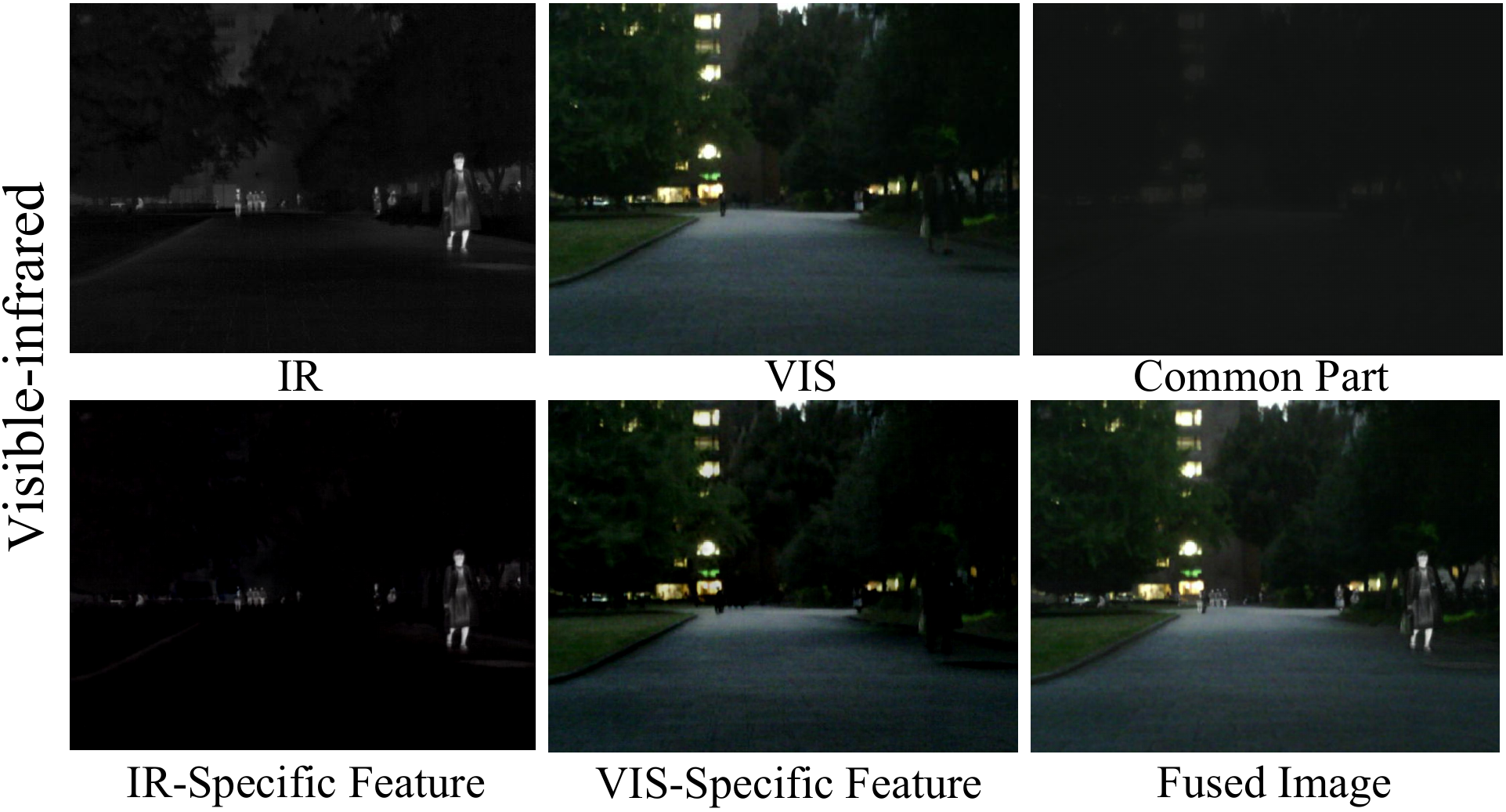}
    
    \vspace{-0.2cm}
	\caption{{ Visual demonstration of the fusion process.}}
    \label{F6}
    \vspace{-0.2cm}
\end{figure}

Next, we  evaluate the effectiveness of DCIAE framework and $\psi$. Initially, we verify whether DCIAE framework can effectively extract the common features and their respective specific  features of the source images.
We present the visual results in Figure \ref{F6}. 
For MFF, we find that the extracted common part appears as a fully blurred image. In VIF, the shared components are minimal due to the significant modality gap between the source images. These  are consistent with our intuition, which can be primarily attributed to $loss_2$ in (\ref{l2}).
Moreover,  the modality specific features of  source images highlight their distinct contributions to the fusion process.  This verifies that our method is not a `black box', but rather a transparent architecture capable of revealing the contributions of each input modality during fusion.

Next, we validate the effectiveness of the truncation function $\psi$ 
 described in equation (\ref{C77}).
\begin{figure}
	\centering
	~\includegraphics[width=0.48\textwidth]{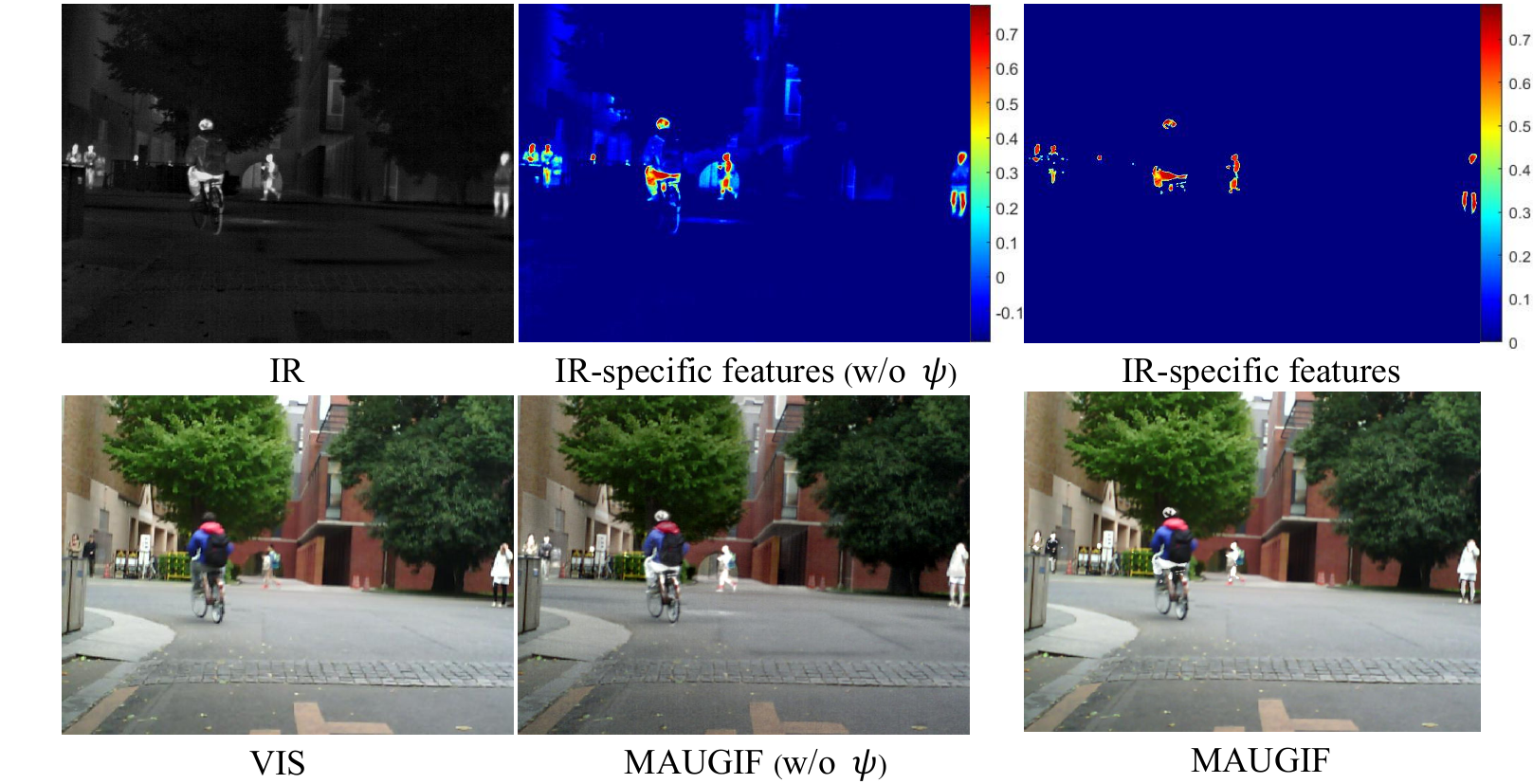}

    \vspace{-0.2cm}
	\caption{{ Ablation study  on $\psi$ in the VIF task.}}
    \vspace{-0.4cm}
    \label{F10}
\end{figure}
 Figure \ref{F10} illustrates that, without $\psi$, the extracted IR-specific features still contain a significant amount of irrelevant environmental information, which is predominantly composed of negative values. This causes the fused image to appear noticeably darker. When $\psi$ is incorporated, the important regions in the IR image are highlighted while less relevant ones are suppressed, resulting in a fused image with virtually no color cast.  This fully validates the effectiveness of $\psi$.

\section{Conclusion}\label{G}
\vspace{-0.2cm}
	In this paper, we propose a mechanism-aware  general image fusion method, termed MAUGIF, which is built upon the DCIAE framework. According to the inherent mechanisms of different fusion tasks,  we introduce a  classification of  additive and multiplicative fusion. Then, we employ the DCIAE framework to decompose source images into common components and modality-specific features.  During the fusion process, the decoders are regarded as  feature injectors  to generate the final fusion images. By aligning the decoder architectures with specific fusion strategies, our design not only supports flexible and effective fusion but also enhances interpretability.  Finally, extensive numerical experiments further validate the effectiveness of our method.

{
    \small
    \bibliographystyle{ieeenat_fullname}
    \bibliography{CiteTex}
}
\end{document}